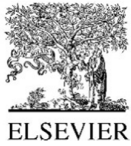

# Improving the Performance of English-Tamil Statistical Machine Translation System using Source-Side Pre-Processing

Anand Kumar M[1], Dhanalakshmi V[2], Soman K P[1] and Sharmiladevi V
[1] CEN, Amrita Vishwa Vidyapeetham, Coimbatore, India
Email: {m_anandkumar,kp_soman}@cb.amrita.edu
[2] Department of Tamil, SRM University, Chennai, India
Email: dhanagiri@gmail.com

*Abstract*—Machine Translation is one of the major oldest and the most active research area in Natural Language Processing. Currently, Statistical Machine Translation (SMT) dominates the Machine Translation research. Statistical Machine Translation is an approach to Machine Translation which uses models to learn translation patterns directly from data, and generalize them to translate a new unseen text. The SMT approach is largely language independent, i.e. the models can be applied to any language pair. Statistical Machine Translation (SMT) attempts to generate translations using statistical methods based on bilingual text corpora. Where such corpora are available, excellent results can be attained translating similar texts, but such corpora are still not available for many language pairs. Statistical Machine Translation systems, in general, have difficulty in handling the morphology on the source or the target side especially for morphologically rich languages. Errors in morphology or syntax in the target language can have severe consequences on meaning of the sentence. They change the grammatical function of words or the understanding of the sentence through the incorrect tense information in verb. Baseline SMT also known as Phrase Based Statistical Machine Translation (PBSMT) system does not use any linguistic information and it only operates on surface word form. Recent researches shown that adding linguistic information helps to improve the accuracy of the translation with less amount of bilingual corpora. Adding linguistic information can be done using the Factored Statistical Machine Translation system through pre-processing steps. And importantly, machine translation system for language pair with disparate morphological structure needs best pre-processing or modeling before translation. English and Tamil languages are belongs to different language family so it is difficult for system to automate the morpho-syntactic mapping between them using statistical methods. This paper investigates about how English side pre-processing is used to improve the accuracy of English-Tamil SMT system.

*Index Terms*— Statistical Machine Translation, Reordering, Linguistic information, Pre-processing, English-Tamil, Morphologically Rich

I. INTRODUCTION

Recently, SMT systems are introduced with linguistic information in order to address the problem of word order and morphological variance between the language pairs. Integrating this linguistic information can be known as pre-processing and this pre-processing of source language is done constantly on the training and testing corpora. More source side pre-processing steps brings the source language sentence closer to that of the target language sentence. Statistical translation models have evolved from the word-based models originally proposed by Brown et.al [1] to syntax-based and phrase-based techniques. The beginnings of phrase-based translation can be seen in the alignment template model introduced by Och et.al [2]. A joint probability model for phrase translation was proposed by Marcu et.al [3]. Koehn et.al [4] used certain heuristics to extract phrases that are consistent with bidirectional word-alignments generated by the IBM models [1]. Phrases extracted using these heuristics are also shown to perform better than syntactically motivated phrases, the joint model, and IBM model-4 [4]. Syntax-based models use parse-tree representations of the sentences in the training data to learn, among other things, tree transformation probabilities. These methods require a parser for the target language and, in some cases, the source language too. Yamada et.al [5] developed a model that transforms target language parse trees to source language strings by applying reordering, insertion, and translation operations at each node of the tree. Various researchers proposed SMT methods based on tree-to-tree mappings. Imamura et.al [6] presented a similar method that achieves notable improvements over a Phrase-based baseline model for Japanese-English translation.

II. RELATED WORKS

As mentioned in the introduction, a lot of pre-processing work has been done on source language side to improve the performance of statistical machine translation. Recent developments showed an improvement in translation quality when using the explicit syntax based reordering. One of these developments is the pre-translation approach which alters the word order of source language sentence to target language word order before translation. This is done based on predefined linguistic rules that are either manually created or automatically learned from parallel corpora. Notable researches in SMT in order to improve the performance of Statistical Machine Translation (SMT) system using morpho-syntactic information is also explained in this section. This section explains the detailed literature review about the existing pre-processing methods for machine translation.

*A. Handling Word-Order Differences*

Various methods for reordering have been developed for handling the word-order difference between language pairs. Reordering is successfully applied for French to English and from German to English translation system. Reordering rules are often used to improve the translation quality, with these reordering rules being automatically learned from the parse trees for both source and target sentences. Marta Ruiz Costa-juss`a 2006, [7] proposed a novel reordering algorithm for SMT. They introduced two new approaches; they are block reordering and Statistical Machine Reordering (SMR). They also explained various reordering methods like syntax based reordering and heuristic reordering in 2009 [8]. Irimia Elena and Alexandru Ceauşu (2010) [9] presented a method for extracting translation examples using the dependency linkage of both the source and target language sentence. They identified two types of dependency link-structures super-links and chains and used these structures to set the translation example borders. Sriram Venkatapathy et.al (2010) [10] proposed a dependency based statistical system that uses discriminative techniques to train its parameters. Experiments are conducted for English- Hindi parallel corpora. Ananthakrishnan R, et.al [11] developed a syntactic and morphological pre-processing for English to Hindi SMT system. They reorder the English source sentence as per Hindi syntax, and segment the suffixes of Hindi for morphological processing.

*B. Handling the Morphological Differences*

Morphological variations between the language pairs have been handled using the different type of methodologies. Few methods are dedicatedly developed to augment the SMT models with morphological information to improve the quality of translation into morphologically rich languages. Soha Sultan (2011) [12] introduced two approaches to augment the linguistic knowledge with English-Arabic statistical machine translation (SMT). The first approach improves SMT by adding linguistically motivated syntactic features to particular phrases. These added features are based on the English syntactic information, namely part-of-speech tags and dependency parse trees. The second approach improves morphological agreement in machine translation output through post-processing. This method uses the projection of the English dependency parse tree onto the Arabic sentence in addition to the Arabic morphological analysis in order to extract the agreement relations between words in the Arabic sentence. Individual morphological features are trained using syntactic and morphological information from both the source and target languages. The predicted morphological features are then used to generate the correct surface forms.

Rabih M. Zbib (2010) [13] presented the methods for using linguistically motivated information to enhance the performance of statistical machine translation (SMT). This pre-processing reduces the gap in the complexity of the morphology between Arabic and English language. Finally the system combines the outputs of an SMT system and a Rule-based MT (RBMT) system, taking advantage of the flexibility of the statistical approach and the rich linguistic knowledge embedded in the rule-based MT system.

Panagiotis (2005) [14] proposed a novel algorithm for incorporating morphological knowledge for English to Greek Statistical Machine Translation (SMT) system. He suggested a method of improving the translation quality of existing SMT systems, by incorporating word-stems into SMT systems. Reyyan Yeniterzi and Kemal Oflazer (2010) [15] presented a novel way to incorporate source syntactic structure in phrase-based machine translation by parsing the source sentences and then encoding many local and nonlocal source syntactic structures as additional complex tag factors. They have developed set of about 20 linguistically motivated syntax-to-morphology transformations for English-Turkish. The limitation of this approach is that it is not directly applicable in the reverse direction.

Loganathan et.al [16] developed suffix-separation rules for English and Tamil languages and evaluates the impact of suffix splitting on translation quality. This paper describes how the developed pre-processing method improves the English-Tamil phrase based and Hierarchal SMT system. He also demonstrated that suffix separation helps in reducing the data sparse problem. Tirumeni et.al (2011) [17] proposed a technique to handle the phrasal verbs and idioms for English to Tamil machine translation. This method identifies phrasal verbs and grouping them in English and its equivalent in Tamil prior to training and testing. Sara Stymne (2009), [18] explored how compound processing can be used to improve the phrase-based statistical machine translation (PBSMT) between English and German/Swedish. For translation into Swedish and German the parts are merged after translation. The effect of different splitting algorithms for translation between English and German, and of different merging algorithms for German is also investigated.

This paper is organized as follows, Section III presents the syntax and morphological comparison between English and Tamil language. Section IV describes the detailed methodology of pre-processing which includes reordering and compounding and Section V illustrates the experiments and results for the Baseline and Factored SMT system. The final section gives the concluding remarks and future possibilities.

### III. MORPHO-SYNTACTIC COMPARISON BETWEEN ENGLISH AND TAMIL LANGUAGE

Grammar of a language is divided into syntax and morphology. Syntax is how words are combined to form a sentence and morphology deals with the formation of words. Morphology is also defined as the study of how meaningful units can be combined to form words. One of the reasons to process a morphology and syntax together in language processing is that a single word in a language is equivalent to combination of words in another. The term "*morpho-syntax*" is a hybrid word that comes from morphology and syntax. Morpho-syntax plays a major role in processing different types of languages and it is also a related term to machine translation because the elementary unit of machine translation is words and phrases. Retrieving the syntactic information is a primary step in translation between any two languages. The tool which is used for retrieving syntactic structure from a given sentence is called parsing and which is used to retrieve morphological features from a word is called as morphological analyzer. Syntactic information includes dependency relation, syntactic structure and POS tags. Morphological information consists of lemma and morphological features. Linguistically a sentence in any language can be analysed using parser and morphological analyzer.

Tamil is an agglutinative and morphologically rich language. Tamil words consist of a lexical root to which one or more affixes are attached. Mostly, Tamil affixes are suffixes. Tamil suffixes can be derivational suffixes, which either changes the Part-of-Speech of the word or its meaning, or inflectional suffixes, which mark categories such as person, number, mood, tense, etc. Tamil POS taggers [19] and Morphological Analyzers [20] [21] are exists with good precision.

#### A. Syntactic Comparison

This sub-section gives a closer look and notable differences between the syntax of English and Tamil language. Syntax is a theory of sentence structure and it guides reordering when translations between a language pair contain disparate sentence structure. English and Tamil are from different language families.

289

English is an Indo-European language and Tamil is a Dravidian language. English has the word order of Subject–Verb-Object (SVO) and Tamil has the word order of Subject-Object-Verb (SOV). For example, the main verb of a Tamil sentence always comes at the end but in English it comes between subject and object. English is a fixed word-order language where Tamil word-order is flexible. Flexibility in word order represent that the order may change freely without affecting the grammatical meaning of the sentence. While translating from English to Tamil, English verbs have to be moved from after the subject to end of the sentence. Fig.1 shows the word-order difference in English and Tamil sentences.

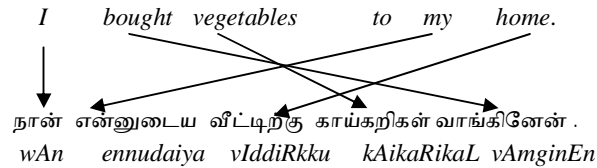

*I    bought   vegetables    to    my    home.*

நான்   என்னுடைய   வீட்டிற்கு   காய்கறிகள்   வாங்கினேன் .
*wAn    ennudaiya    vIddiRkku    kAikaRikaL    vAmginEn*

Figure 1. Word-order difference in English and Tamil.

*B. Morphological Comparison*

Morphology is the study of structure of words in a language. Words are made up of morphemes. These are the smallest meaningful unit in a word. For example, "pens" is made of "pen" + "s", here "s" is a plural marker, "talked" is made of "talk" + "ed" , here "ed" represents past tense. English is morphologically simple language but Tamil is a morphologically rich language. Morphology is one of the significant terms for improving the performance of machine translation system.

Morphological difference between English and Tamil complicates the Statistical Machine Translation task. English language mostly conveys the relationship between words using function words or location of the words but Tamil language expresses using morphological variations of word. Therefore, Tamil language had larger vocabulary of surface forms. This led to sparse data problem in English to Tamil SMT system. In order to solve this, large amount of parallel training corpora is required to cover the entire Tamil surface form. It is very difficult to create or collect the parallel corpora which contain all the Tamil surface forms because Tamil is one of the less resourced languages. Instead of covering entire surface forms a new method is required to handle all word forms with the help of limited amount of data. Factored SMT is a suitable model for morphologically rich language like Tamil. Generally, languages not only differ in the word order but also differ in encoding the relationship between words. English language is strictly in fixed word order and involves heavy usage of function words but less usage in morphology. Tamil language had a rich morphological structure and heavy usage of content word but free word-order language. Because of the function words, the average number of words in English sentences is more when compared to the words in an equivalent Tamil sentence.

Tamil translations of English function words do not independently exist because these words are coupled with Tamil content words and this leads to alignment and sparse data problem. English language contains more function words than content words but Tamil language has more content words. Table I shows the various word forms based on English tenses. In Tamil, verbs are morphologically inflected due to tense and PNG (Person-Number-Gender) markers and nouns are inflected due to count and cases. Each Tamil verb root is inflected into more than ten thousand surface word forms because of agglutinative nature of Tamil language [22]. This morphological richness of Tamil language leads to sparse data problem in Statistical Machine Translation system. Examples of Tamil word forms based on tenses are given in Table II.

TABLE I – ENGLISH TENSES AND WORD-FORMS

| Root Word | Tenses | Word Form |
|---|---|---|
| Play | Simple Present | Play |
| | Present Continuous | is playing |
| | Present Perfect | have played |
| | Past | Played |
| | Past perfect | had played |
| | Future | will play |
| | Future Perfect | will have played |

TABLE II – TAMIL TENSES AND WORD-FORMS

| Root | Tenses | Word Form |
|---|---|---|
| விளையாடு *(vilayAdu)* | Present+1S | விளையாடுகின்றேன் *vilayAdu-kinR-En* |
| | Present+3SN | விளையாடுகின்றது *vilayAdu-kinR-athu* |
| | Present+3PN | விளையாடுகின்றன *vilayAd-kinR-ana* |
| | Past+1S | விளையாடினேன் *vilayAd-in-En* |
| | Past+3SM | விளையாடினான் *vilayAd-in-An* |
| | Future+2S | விளையாடுவாய் *vilayAdu-v-Ay* |
| | Future+3SF | விளையாடுவாள் *vilayAdu-v-AL* |

IV. METHODOLOGY FOR PRE-PROCESSING SOURCE SENTENCES

This section explains the pre-processing methods for English sentence to improve the quality of English to Tamil Statistical Machine Translation system. The pre-processing module for English language sentence includes three stages, which are reordering, factorization and compounding. Fig.2 shows the pre-processing stages of English language sentence. The first step in pre-processing English sentence is to retrieve the linguistic features such as lemma, POS tag, and syntactic relations using Stanford parser Tool [23].

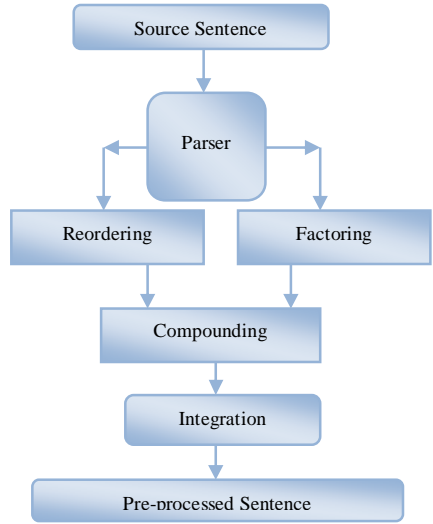

Figure 2. Framework for Pre-processing.

These linguistic features along with the sentence will be subjected to reordering and factorization stages. Reordering applies the manually-created reordering rules to the syntactic trees for rearranging the phrases in the English sentences. Factorization takes the surface words in the sentence and then factored using syntactic tool. This information is appended to the words in the sentence. Part-of-Speech tags are simplified and included as a factor in factorization. This factored sentence is given to the compounding stage. Compounding is defined as adding additional morphological information to the morphological factor of source (English) language words. Additional morphological information includes function word, subject information, dependency relations, auxiliary verbs, and model verbs. This information is based on the morphological structure of the target language. After adding this information, few function words and/or auxiliary information are removed and reordered information is incorporated in integration phase

*A. Reordering the word-order in source language Sentences*

Reordering transforms the source language sentence into a word order that is closer to that of the target language. Mostly in Machine Translation system the order of the words in the source language sentence is often different from the words in the target language sentence. The word-order difference between source and target languages is one of the most significant errors in a Machine Translation system. Phrase based SMT systems are limited for handling long distance reordering. A set of 180 syntactic reordering rules are developed and applied on the English language sentence to better align with the Tamil sentence. Sample reordering rules are shown in Table III and these rules elaborate the structural differences of English and Tamil sentences. These transformation rules are applied to the parse trees of the English source language. Parse trees are generated using Stanford parser tool [23]. Quality of parse trees plays an important role in syntactic reordering.

In this paper, the source language is English and therefore the parses are more accurate and the reordering based on the parses are exactly matched with target language. Generally, English parsers are performing better than other language parsers because, English parsers developed from longer and advanced statistical parsing techniques are applied. Fig.3 shows the methodology for reordering. Production rules are retrieved from the syntax tree. The syntactic tree for the example sentence is shown in Fig.4.

TABLE III – SAMPLE REORDERING RULES

| Source | Target |
|---|---|
| S -> NP VP | # S -> NP VP |
| PP -> TO NP-PRP | # PP -> TO NP-PRP |
| VP -> VB NP* SBAR | # VP -> NP* VB SBAR |
| VP -> VBD NP | # VP -> NP VBD |
| VP -> VBD NP-PP | # VP -> PP-NP VBD |
| VP -> VBP PP | # VP -> PP VBP |

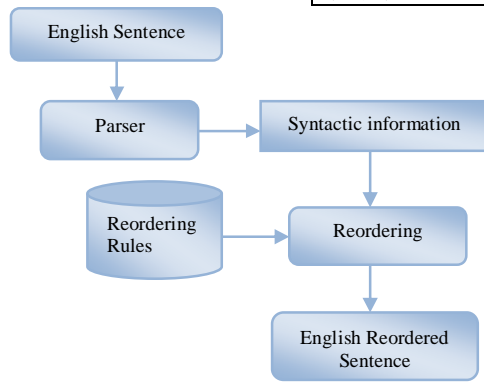

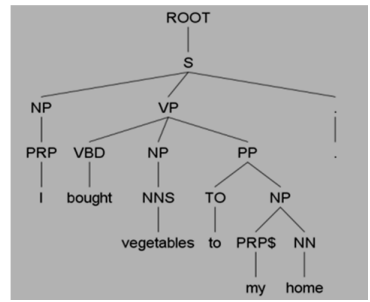

Figure 3. Process of Reordering.      Figure 4. Sample English Syntactic Tree

For instance, take an example reordering rule.
    VP -> VBP PP# VP -> PP VBP# 0:1,1:0

Reordering rules consists of three units.
 i. Production rules of original English sentence (source).
 ii. Transformed production rules according to Tamil sentence (target).
iii. Source part numbers and target part numbers. These numbers indicate the reorder of the source sentence (transformations).

Where, # divides the units of reordering rules, the last unit indicates source and target indexes. In the above example, "0:1, 1:0" indicates first child of the target rule is from second child of the source rule; second child of the target rule is from first child of the source rule.

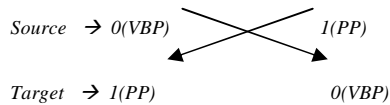

*Source* → *0(VBP)*    *1(PP)*

*Target* → *1(PP)*    *0(VBP)*

For example take an English sentence, "*I bought vegetables to my home*".
*Production rules of the English Sentence:*
    i. S->NP VP
    ii. VP->VBD NP PP
    iii. PP->TO NP
    iv. NP->PRP$ NN

The first production rule (i) S->NP VP is matched with the first reordering rule in Table III. The target transformation is same as the source pattern and therefore no change in first production rule. The next production rule (ii) VP->VBD NP PP is matched with the fifth reordering rule in the table and the transformation is 0:2  1:1  2:0 , it means that source word order (0,1,2) is transformed into (2,1,0). (0,1,2) are the index of VBD NP and PP, now the transformed pattern is PP NP VBD. This process is continuously applied to each of the production rules. Finally the transformed production rule is given below.

*Reordered Production rules of the English sentence:*
   i. S->NP VP
  ii. VP->*PP  NP VBD*
 iii. PP->*NP  TO*
  iv. NP->*NN  PRP$*

TABLE IV - ORIGINAL AND REORDERED SENTENCES

| Original Sentences | Reordered Sentences |
|---|---|
| I saw a beautiful child | I a beautiful child saw |
| He came last week | He last week came |
| Sharmi gave her book to Arthi | Sharmi her book Arthi to gave |
| She went to shop for buying fruits | She fruits buying for shop to went |
| Cat is sleeping on the table | Cat the table on sleeping is. |

Using this Reordered production rules the English Sentence is re-generated and the sentence is, *"I my home to vegetables bought"*. English parallel corpora which is used for training is reordered and the testing sentences are also reordered. 80% of English sentences are reordered correctly according to the rules which are developed. Examples of original and reordered English sentences are shown in Table IV. After reordering the English sentences are subjected to the compounding stage.

*B. Factoring the source language Sentences*

English factorization is considered as one of the important pre-processing step. Factorization splits the surface word into linguistic factors and integrates as a vector. Instead of mapping surface words in translation, factored models maps the linguistic units (factors) of language pair. Stanford Parser is used for factorizing English language sentence. From the parser output, linguistic information such as, lemma, part-of-speech tags, syntactic information and dependency information are retrieved. This linguistic information is integrated as factors into the surface word.

The current phrase-based models are limited to the mapping of small text chunks without the use of any explicit linguistic information like morphological and syntactical. Such information plays a significant role in morphologically rich languages. In other hand, for many language pairs, the availability of bilingual corpora is very less. SMT performance is based on the quality and quantity of corpora. So, SMT strictly needs a new method which uses linguistic information explicitly with fewer amounts of parallel data. Factored translation framework for statistical translation models to tightly integrate linguistic information. It is an extension of phrase-based Statistical Machine Translation that allows the integration of additional morphological and lexical information, such as lemma, word class, gender, number, etc., at the word level on both source and the target languages. Factorization of parallel sentences is a fundamental step in factored machine translation system. Tamil factors are retrieved using the existing Tamil POS Tagger [19] and Morphological analyzer systems [20]. Factored translation model is one way of representing linguistic knowledge to Statistical machine translation explicitly. Factors which are considered in pre-processing and their description of English language are shown in Table V.

In this example, *word* refers surface word, l*emma* represents the dictionary word or root word, *word class* represents word-class category and *morphology* factor represents a compound-tag which contains morphological information and/or function words. In some cases the "*morphology*" factor, also contains the dependency relations and/or PNG information. For instance, the English sentence, "*I bought vegetables to my home*", is factored into linguistic factors which are shown in Table VI.

*C. Compounding the morphology information in Source Language Sentence*

A baseline Statistical Machine Translation (SMT) system only considers surface word forms and does not use linguistic information. Translating into target surface word form is not only dependent on the source word-form and I t also depends on additional morpho-syntactic information. While translating from morphologically simpler language to morphological rich language, it is very hard to retrieve the required morphological knowledge from the source language sentence. This morphological information is an important term for producing a target language word-form. This preprocessing phase compounding is used to retrieve the required linguistic information from source language sentence for generating target word. Morphologically rich languages have a large number of surface forms in the lexicon to compensate for a free word-order. This large number of word-forms in Tamil language is very difficult to generate from English language words. Fig.5 illustrates the process of compounding methodology.

TABLE V - DESCRIPTION OF THE FACTORS IN ENGLISH WORD

| FACTORS | DESCRIPTION | EXAMPLE |
| --- | --- | --- |
| Word | Surface words or word forms | Coming, went, beautiful, eyes |
| Lemma | Root word or Dictionary word | Play, run, home, pen |
| Word Class | Minimized POS tag | N, V, ADJ, ADV |
| Morphology | POS tag, dependency information, function words, subject information, Auxilary and model verbs | VBD, NNS, nsubj, pobj, to, has been, will |

TABLE VI - EXAMPLE OF ENGLISH WORD FACTORS

| WORD | FACTORS |
| --- | --- |
| I | i\|PRP\|PRP_nsubj |
| bought | buy\|V\|VBD |
| vegetables | vegetable\|N\|NNS_dobj |
| to | to\|PRE\|TO_prep |
| my | my\|PRP\|PRP$_poss |
| home | home\|N\|NN_pobj |

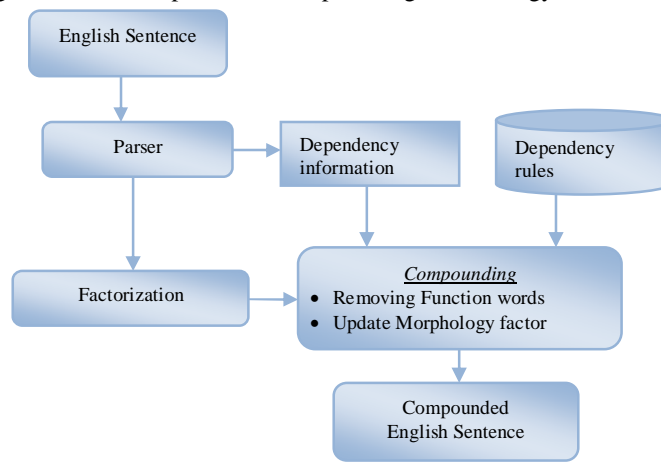

Figure 5. Compounding methodology.

Compounding add the extra morphological information to the morphological factor of source (English) language words. Additional morphological information includes subject information, dependency relations, auxiliary verbs, model verbs and few function words. This information is based on the morphological structure of Tamil language. In compounding phase, dependency relations are used to identify the function words from the English factored corpora. During integration, few function words are deleted from the factored sentence and attached as a morphological factor to the corresponding content word. In Tamil language, function words are not directly available but it is fused with corresponding content word. So instead of making the sentences into similar representation, function words are removed from an English sentence. This process reduces the length of the English sentences. Like function words, auxiliary verbs and model verbs are also identified and attached in morphological factor of head word of source sentence. Now the morphological factor representation of the English language sentence is similar to that of the Tamil language sentence. This compounding step indirectly integrates dependency information into the source language factor.

Compounding also identifies the subject information from English dependency relations. This subject information is folded into the morphological factor of English verb and it helps to identify the PNG (Person-Number-Gender) marker for Tamil language during translation. PNG marker plays an important role in Tamil morphology due to the subject-verb agreement nature of Tamil language. Most of the Tamil verbs are generated using this PNG marker. English auxiliary verbs are also identified from the dependency information and then removed and folded in morphological factor of the head word/verb. English sentence is factorized and then subjected to the compounding phase. A word in factorized sentence includes part of speech and morphological information as factors. Compounding takes dependency relations from Stanford parser and produces the compounded sentences using pre-defined linguistic rules. These rules are developed based on morphological difference between English and Tamil language. This rule identifies the transformations from English morphological factors to Tamil morphological factors.

## V. EXPERIMENTS AND RESULTS

Four experiments are performed to evaluate the impact of source-side pre-processing in SMT system. The first experiment, Baseline, is performed with Phrase based SMT system and the other three experiments are based on Factored SMT models. In basic factored model, the target surface forms are generated using the in-build Generation model and another factored model used the existing Tamil morphological generator system [22] for generating the same. The final experiment is the combination of pre-processing and Tamil morphological generator in Factored SMT. The size of 10,000 English-Tamil parallel sentences in tourism domain from EILMT project (English Indian Languages Machine Translation System, funded by TDIL program, DeitY) is used in our experiments. From this corpus 9000 sentences are used as training set and 1000 sentences as testing set. Development set does not improve the performance of factored SMT system so tuning part is omitted in these experiments.

Automatic translation metrics are most commonly measured by comparing the translation output to reference translation and providing some kind of score. There are huge number of automatic metrics are exist but the well-known metrics BLEU [24] and METEOR [25] are focused in this paper. BLEU scores are obtained in terms of BLEU unigram, BLEU 4-gram and cumulative BLEU. These accuracies are shown in Table VII. This table reveals that using source-side pre-processing in SMT increased the BLEU score by 56% in cumulative BLEU, 20% in 1-gram. The Table 8 illustrates the lemma-wise BLEU scores of the developed SMT system on the test sets. Due to the mistakes in Tamil morphological generator and morphological translation in SMT, BLEU scores for surface-form is less compared to the lemma-wise results. METEOR scores based on surface word form and lemma are calculated and shown in Table IX. This table describes that using source-side pre-processing in SMT increased the METEOR score by 54% for surface forms and 38% for lemmas. As Table VII, VIII and IX demonstrates, both reordering and compounding method as such have a positive effect on the translation quality. The results revealed that the preprocessing source-side sentences according to target language improve the translation quality significantly. Fig.6 shows the comparison of BLEU scores in lemma-wise and surface form based results.

TABLE VII – BLEU SCORES FOR SMT BASED ON SURFACE FORMS

| SYSTEM | BLEU | BLEU-1 | BLEU-4 |
|---|---|---|---|
| BASELINE | 2.92 | 24.4 | 0.4 |
| FACT | 2.20 | 18.31 | 0.2 |
| FACT+ MORPHGEN | 3.4 | 26.9 | 0.6 |
| FACT+RR+COMP+MORPHGEN | 5.32 | 32.10 | 1.30 |

TABLE VIII – LEMMA-WISE BLEU SCORES FOR F-SMT

| SYSTEM | BLEU | BLEU-1 | BLEU-4 |
|---|---|---|---|
| FACT+ MORPHGEN | 9.21 | 38.39 | 2.40 |
| FACT+RR+COMP+MORPHGEN | 12.51 | 41.00 | 4.60 |

TABLE IX – METEOR SCORES FOR SMT

| System | Meteor | MetEor $_{LEMMA}$ |
|---|---|---|
| BaseLine | 0.11 | - |
| Fact | 0.091 | - |
| Fact+ MorphGen | 0.136 | 0.248 |
| Fact+RR+Comp+MorphGen | 0.211 | 0.342 |

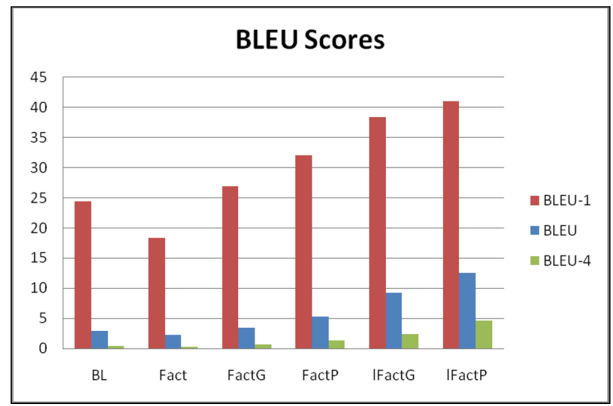

Figure 6. Comparison of BLEU scores.

VI. CONCLUSIONS

This paper presents the impact of source-side linguistic pre-processing in English-Tamil SMT system. Pre-processing stages includes reordering, factoring and compounding and these techniques are matches the English language sentence with Tamil language sentence. Finally integration process incorporates the pre-processing stages. The paper has also presented the syntactic and morphological variance between English and Tamil language. However, reordering plays an important role especially for language pairs with disparate sentence structure. The difference in word order between two languages is one of the most significant sources of errors in Machine Translation. While phrase based MT systems do very well at reordering inside short windows of words, long-distance reordering seems to be a challenging task. The translation accuracy can be significantly improved if the reordering is done prior to translation. Compounding and factoring are used in order to reduce the amount of English-Tamil bilingual data. Pre-processing also reduces the number of words in English sentence. Accuracy of pre-processing heavily depends on the quality of the parser. Different researches have proven that pre-processing is the effective method in order to obtain a word-order and morphological information which match the target language. Moreover, this pre-processing approach can be especially applicable for translating sentences from morphologically simple languages to morphologically rich languages. Reordering rules and compounding rules which are proposed in this paper can be used for other Dravidian languages with small modifications. In future, automatic rule extraction for reordering and compounding using bi-lingual corpora will improve the accuracy of machine translation system and also this automation can be common method for any language pairs. This results obtained from the experiments has proved that adding linguistic knowledge in pre-processing of training data can lead to remarkable improvements in translation performance.